\documentclass[conference,letterpaper]{IEEEtran}

\usepackage[utf8]{inputenc} 
\usepackage[T1]{fontenc}
\usepackage{url}
\usepackage{ifthen}
\usepackage{cite}
\usepackage[cmex10]{amsmath} % Use the [cmex10] option to ensure complicance

\usepackage{bm,mathtools,amsfonts,amssymb,bbm,float,hyperref,euscript,setspace,tikz,steinmetz,cite,enumitem,wrapfig,algpseudocode,subfigure,enumitem}

\usepackage{lastpage}

\usepackage{algorithm}
\usepackage{algorithmicx}
\usepackage{algpseudocode}

\usepackage[normalem]{ulem}
\usepackage[mode=buildnew]{standalone}
\usepackage{mathabx} % for convolution
\newtheorem{theorem}{Theorem}

\usepackage{mleftright}\mleftright
\newcommand*{\medcap}{\mathbin{\scalebox{1.5}{\ensuremath{\cap}}}}%

\definecolor{darkgreen}{rgb}{0, 0.5, 0}
\definecolor{darkred}{RGB}{128, 0, 0}
\newcommand{\stkout}[1]{\ifmmode\text{\sout{\ensuremath{#1}}}\else\sout{#1}\fi}

\newcommand{\vc}[1]{\mathbf{#1}} % vector
\newcommand{\der}{\mathrm{d}} % 

\newcommand{\eg}{\emph{e.g., }}
\newcommand{\ie}{\emph{i.e., }}

\DeclareMathOperator{\gen}{gen}

\DeclareMathOperator*{\argmax}{arg\,max}
\DeclareMathOperator*{\argmin}{arg\,min}

\makeatletter
\DeclareFontFamily{OMX}{MnSymbolE}{}
\DeclareSymbolFont{MnLargeSymbols}{OMX}{MnSymbolE}{m}{n}
\SetSymbolFont{MnLargeSymbols}{bold}{OMX}{MnSymbolE}{b}{n}
\DeclareFontShape{OMX}{MnSymbolE}{m}{n}{
	<-6>  MnSymbolE5
	<6-7>  MnSymbolE6
	<7-8>  MnSymbolE7
	<8-9>  MnSymbolE8
	<9-10> MnSymbolE9
	<10-12> MnSymbolE10
	<12->   MnSymbolE12
}{}
\DeclareFontShape{OMX}{MnSymbolE}{b}{n}{
	<-6>  MnSymbolE-Bold5
	<6-7>  MnSymbolE-Bold6
	<7-8>  MnSymbolE-Bold7
	<8-9>  MnSymbolE-Bold8
	<9-10> MnSymbolE-Bold9
	<10-12> MnSymbolE-Bold10
	<12->   MnSymbolE-Bold12
}{}

\let\llangle\@undefined
\let\rrangle\@undefined
\DeclareMathDelimiter{\llangle}{\mathopen}%
{MnLargeSymbols}{'164}{MnLargeSymbols}{'164}
\DeclareMathDelimiter{\rrangle}{\mathclose}%
{MnLargeSymbols}{'171}{MnLargeSymbols}{'171}
\makeatother

\begin{document}
\fontencoding{OT1}\fontsize{10}{11.4}\selectfont

\title{Minimal Communication-Cost Statistical Learning}

%%% Several authors with up to three affiliations:
\author{%
  \IEEEauthorblockN{Milad Sefidgaran$^{\:\nmid}$ \qquad \qquad Abdellatif Zaidi$\:^{\dagger}$$\:^{\nmid}$ \qquad \qquad Piotr Krasnowski$^{\:\nmid}$}
  \IEEEauthorblockA{$^{\:\nmid}$ Paris Research Center, Huawei Technologies France \qquad \qquad
  $\:^{\dagger}$ Universit\'e Gustave Eiffel, France\\
  \texttt{\{milad.sefidgaran2,piotr.g.krasnowski\}@huawei.com, abdellatif.zaidi@univ-eiffel.fr}}
}

\maketitle

\begin{abstract}
A client device which has access to $n$ training data samples needs to obtain a statistical hypothesis or model $W$ and then to send it to a remote server. The client and the server devices share some common randomness sequence as well as a prior on the hypothesis space. In this problem a suitable hypothesis or model $W$ should meet two distinct design criteria simultaneously: (i) small (population) risk during the inference phase and (ii) small `complexity' for it to be conveyed to the server with minimum communication cost.  In this paper, we propose a joint training and source coding scheme with provable in-expectation guarantees, where the expectation is over the encoder's output message. Specifically,  we show that by imposing a constraint on a suitable Kullback-Leibler divergence between the conditional distribution induced by a compressed learning model $\widehat{W}$ given $W$ and the prior, one guarantees simultaneously small average empirical risk (aka training loss), small average generalization error and small average communication cost. We also consider a one-shot scenario in which the guarantees on the empirical risk and generalization error are obtained for every encoder's output message.
\end{abstract}

\section{Introduction}
The development of distributed or decentralized machine learning solutions has witnessed a rapid increase over the last years, in particular, due to abundant applications in various areas. Examples include the federated learning (FL) of~\cite{mcmahan2017}, the split learning (SL) of~\cite{gupta2018distributed} and the in-network learning (INL) of~\cite{aguerri21,moldoveanu2021network}. Often,  the client devices in these architectures (also simply referred to as ``clients") process their available training data samples locally and then send the output hypotheses or models to a central node or server. The process is repeated until a given loss function is minimized over the training data set; and, typically, this induces a large communication overhead. For this reason, the search for efficient model compression techniques is of paramount importance, especially in bandwidth-constrained settings such as model transmission over a finite-capacity wireless channel.  

Existing approaches to model training and compression are mostly based on a ``separation'' principle. That is, a first processing stage during which the client learns a suitable (local) model or hypothesis on the basis of the available training dataset followed by an independent second processing stage during which the client produces a compressed version of the obtained model that it conveys to the server.  Specifically, there exist two main techniques for model training and transmission:

\textit{Model (update) compression:} In this class of methods, the client first learns a model using its available dataset. Then, it compresses the obtained model (or its update) using techniques such as \eg quantization \cite{dettmers20158,alistarh2017qsgd,bernstein2018signsgd,bernstein2018signsgdII,wu2018error,li2018network,yu2019exploring,jastrzebski2021catastrophic}, sparsification \cite{stich2018sparsified,aji2017sparse,alistarh2018convergence} or combination of them  \cite{basu2019qsparse,dryden2016communication,strom2015scalable}, as well as ``Low-rank decomposition'' \cite{arora2018stronger,li2018algorithmic,jain2016streaming}. The reader is referred to \cite{xu2020compressed} for more details on this class of methods.

\textit{Codebook-based compression:} This class of methods was initiated by \cite{havasi2018minimal} which uses~\cite{harsha2007communication}, and it is based on the aforementioned ``separation'' principle. Here, the client and the server first use a shared prior $Q$ and randomness $\vc{U}$ to agree on a common model source codebook $\mathcal{C}=\{W_1,W_2,...,W_N\}$; and, then, the client sends a locally trained model $W$ by choosing a suitable associated index $i$ in the source codebook, \eg via a variant of \emph{importance sampling} \cite{theis2022algorithms,isik2023communication}. Upon receiving the index $i$, the server uses the codebook $\mathcal{C}$ to recover the model $W_i$. It is shown experimentally that this method can reduce bandwidth consumption up to 50 times compared with the classical model compression methods~\cite{isik2023communication}.  

In all aforementioned prior art works, the problem of model transmission is studied \textit{irrespective} of how well the conveyed hypothesis or model performs during the inference phase, i.e., the population risk. For instance, the sent compressed model is only guaranteed to perform well on the training dataset. In this paper, we consider the problem of joint design of model training and compression in a manner that guarantees simultaneously good performance during the inference phase and minimal communication cost. In doing so, we constraint the source encoder not to know the \emph{learning distribution}, \ie the conditional $P_{W|S}$ induced by the learning algorithm where $W$ is the chosen model or hypothesis and $S$ is the available training dataset. For instance, we depart from previous analyses, such as \cite{isik2023communication}, in two main aspects: \textbf{i)} They analyzed only the communication performance (rate versus distortion), while we analyze the communication rate jointly with the generalization error and the empirical risk of the recovered model,  and \textbf{ii)} our analysis does not require knowledge of the often difficult-to-estimate conditional $P_{W|S}$, especially in high-dimension statistical learning problems. A question similar to that of \cite{isik2023communication} has also been studied in \cite{pase2023semantic} for the problem of sending $n$ ``concepts'' (or equivalently stochastic mappings). In both \cite{isik2023communication} and \cite{pase2023semantic}, the studied question is about sending (a possibly distorted version of) the \emph{learned} model. In other words, two problems of learning and source encoding are studied \emph{separately}. Besides, it is assumed in \cite{pase2023semantic} that the communication constraints can only negatively affect performance, see e.g. the discussion after \cite[Definition~IV.3.]{pase2023semantic}.

\textbf{Contributions.} In this paper, we propose a joint training and source coding scheme whose analysis reveals that by imposing a constraint on a suitable Kullback-Leibler divergence between the conditional distribution induced by a compressed learning model $\widehat{W}$ given $W$ and the prior, one can provably guarantee simultaneously small average training loss, small average generalization error, and small average communication cost. In part, the proof techniques use and extend judiciously those of~\cite{Sefidgaran2022,sefidgaran2023datadependent}, which established formal connections between the generalization error of a statistical learning algorithm and its ``compressibility'' in a suitable information-theoretic sense. Furthermore, we also consider a one-shot variant of the scheme, that we analyze, in which the guarantees on the empirical risk and generalization error are shown to hold for every single encoder output message.

\vspace{-0.2 cm}
\section{Problem setup}
\begin{figure}
    \centering
    \includegraphics[width=0.9\linewidth]{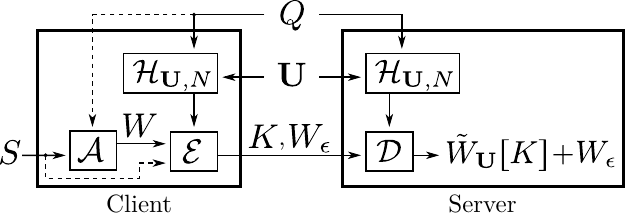}
    \caption{Considered setup for joint local training and remote source coding}
    \label{fig:fig1}
\end{figure}

We consider a point-to-point setup for joint local training and remote source coding, as depicted in Fig.~\ref{fig:fig1}. 

\textbf{Data.} Let $Z$ be the input data taking values over the input space $\mathcal{Z}$, according to an unknown distribution $\mu$. We assume a training dataset $S\coloneqq \{Z_1,\ldots,Z_n\}\sim \mu^{\otimes n}\eqqcolon P_S$ is available.

\textbf{Learning algorithm.} In this work, we consider a general stochastic learning framework. Suppose that the learning algorithm $\mathcal{A}\colon \mathcal{Z}^n \to \mathcal{W}$, by having access to $S$, picks a hypothesis $\mathcal{A}(S)=W\in \mathcal{W}$, possibly non-deterministically. Here, $\mathcal{W}\in \mathbb{R}^d$ denotes the \emph{hypothesis space}.  We denote the conditional distribution induced by this learning algorithm by $P_{W|S}$, and the joint distribution of $(S,W)$ by $P_{S,W}$. Furthermore, denote the marginal distribution of $W$ under $P_{S,W}$ by $P_W$. An example of such a learning algorithm is the popular SGD algorithm.

\textbf{Loss function and risks.} The quality of the prediction of a model $w\in \mathcal{W}$ is assessed using a loss function $\ell \colon \mathcal{W} \times \mathcal{Z} \to \mathbb{R}_+$. In this work, for simplicity, we assume $\ell(z,w) \in [0,1]$. We denote the population risk with respect to this loss function by $\mathcal{L}(w)\coloneqq \mathbb{E}_{Z\sim \mu}[\ell(Z,w)]$, and the empirical risk by $\mathcal{\hat{L}}(s,w)\coloneqq \frac{1}{n} \sum_{i\in[n]} \ell(z_i,w)$ where we used the short-hand notation $[n]$ for the set $\{1,\ldots,n\}\subset \mathbb{N}^*$. Finally, the generalization error is defined as $\gen(s,w) \coloneqq \mathcal{L}(w) - \mathcal{\hat{L}}(s,w)$.

\textbf{Source codebook generation.} Similarly to \cite{havasi2018minimal,theis2022algorithms,isik2023communication}, we assume that a common \emph{source coding codebook} can be constructed at both the sending and the receiving sides using only a shared prior and common randomness. The prior can be defined over $\mathcal{W}$ or more generally, over a quantized set $\hat{\mathcal{W}} \subseteq \mathcal{W}$. This common source codebook can be used for sending the model. 

Formally, fix some set $\hat{\mathcal{W}} \subseteq \mathcal{W}$ and let $Q\in \mathcal{Q}$ be a prior on $\hat{\mathcal{W}}$. In addition, let $\vc{U}\coloneqq \{U_1,\ldots,U_N\}\in \mathcal{U}^N$, $N\in \mathbb{N}^*$, be the common shared randomness, where $U_i$ are distributed i.i.d. and independent of all other variables. Next, let
	\begin{align}
		\mathcal{H}_{\vc{U},N} \coloneqq \left\{\tilde{W}_{\vc{U}}[1],\ldots,\tilde{W}_{\vc{U}}[N] \right\} \subseteq \hat{\mathcal{W}}^N,    
	\end{align}
be the common codebook of hypotheses, where each $\tilde{W}_{\vc{U}}[j]\sim Q$, for $j\in [N]$, is an instance drawn from the distribution $Q$ using the randomness $U_j$. We assume that $\tilde{W}_{\vc{U}}[j]$ is a deterministic function of $Q$ and $U_j$ in order to enable the client and the server to agree on a common codebook.
	
\textbf{Joint learning and source coding.} Suppose that given the learning algorithm $\mathcal{A}$ and the codebook $\mathcal{H}_{\vc{U},N}$, a (potentially stochastic) source encoder 
    \begin{align}
	\mathcal{E}\colon \mathcal{Z}^n \times \mathcal{W} \to [N],    
    \end{align}
    chooses an index $K=\mathcal{E}(S,\mathcal{A}(S))\in[N]$. This potentially stochastic choice can depend, among others, on $Q$ and also on $P_{W|S}$ (if known). For example in the  Minimum Random Coding (MRC) \cite{havasi2018minimal,isik2023communication} (known also as importance sampling), $\hat{\mathcal{W}}=\mathcal{W}$ and the probability of choosing index $j\in[N]$ is proportional to $\frac{\der P_{W|S}}{\der Q}(\tilde{W}_{\vc{U}}[j])$. However, in practice (\eg in SGD), the induced conditional distribution $P_{W|S}$ is not known. Hence, in this work, in contrast to \cite{havasi2018minimal,theis2022algorithms,isik2023communication}, we assume that the encoder cannot explicitly depend on $P_{W|S}$.

Overall, given $\mathcal{A}$ and $Q$, the  joint learning and source coding algorithm, by taking $S$ and the generated codebook $\mathcal{H}_{\vc{U},N}$, chooses the model 
    \begin{align}
        \tilde{W}_{\vc{U}}\left[K\right]=\tilde{W}_{\vc{U}}\left[\mathcal{E}(S,\mathcal{A}(S))\right].    
    \end{align}

In our framework, in addition to the chosen index $K$, we allow sending further \emph{precision} $W_{\epsilon}\in \mathcal{W}$. This further precision together with $\tilde{W}_{\vc{U}}\left[K\right]$, can be alternatively seen as source coding using a more ``refined codebook''. In case when $W_{\epsilon}=W- \tilde{W}_{\vc{U}}\left[K\right]$, $W_{\epsilon}$ is called ``full precision'' model and when $W_{\epsilon}=\vc{0}$, it is called ``no precision'' model. In practice, $W_{\epsilon}$ is a \emph{quantized} version of the difference $W- \tilde{W}_{\vc{U}}\left[K\right]$. For simplicity, we always assume that given $(K,W,\vc{U})$, $W_{\epsilon}$ is chosen deterministically, $(\tilde{W}_{\vc{U}}\left[K\right]+W_{\epsilon})\in \mathcal{W}$, and $\|W_{\epsilon}\| \leq \big\| W- \tilde{W}_{\vc{U}}\left[K\right] \big\|$.

The server, by having access to the codebook $\mathcal{H}_{\vc{U},N}$ and by receiving the index $K\in [N]$ and the further precision $W_{\epsilon}\in \mathcal{W}$, uses the decoder $\mathcal{D}\colon [N] \times \mathcal{W} \to \mathcal{W}$ with the following simple rule
\begin{align}
    \mathcal{D}(K,W_{\epsilon}) = \tilde{W}_{\vc{U}}\left[K\right]+W_{\epsilon}.
\end{align}
Note that for ease of notation, the dependencies of $\mathcal{E}$ and $\mathcal{D}$ on the codebook $\mathcal{H}_{\vc{U},N}$ are dropped in the notations. We further assume that for all $(W,\vc{U})$,
\begin{align}
\Delta_{\vc{U}}\big(W,K\big)\coloneqq &  \big\|W- \tilde{W}_{\vc{U}}[K]\big\|-\big\|W- \mathcal{D}(K,W_{\epsilon})\big\| , \nonumber
\end{align}
is non-negative and increases with $\|W_{\epsilon}\|$. The rationale behind this assumption is that in a proper source coding scheme, the larger $\|W_{\epsilon}\|$, the refiner should be the decoded codeword $\mathcal{D}(K,W_{\epsilon})$, or in other words, the closer $\mathcal{D}(K,W_{\epsilon})$ should be to $W$. Therefore, the second term decreases as $\|W_{\epsilon}\|$ increases, which makes $\Delta_{\vc{U}}$ increase with $\|W_{\epsilon}\|$.

The recovered model $\mathcal{D}(K,W_{\epsilon})$ depends on both the learning algorithm $\mathcal{A}(S)$ and the encoder $\mathcal{E}(S,W)$, as well as the sent further precision $W_{\epsilon}$. In this work, we propose two $P_{W|S}$-agnostic encoding schemes and, for each of them, we analyze both the performance of the model $\mathcal{D}(K,W_{\epsilon})$ and the required communication cost. More precisely, we analyze jointly
\begin{itemize}[leftmargin=0.6 cm]
    \item[i.] the empirical risk of the model $\mathcal{D}(K,W_{\epsilon})$,
       \item[ii.] the generalization error of the model  $\mathcal{D}(K,W_{\epsilon})$,
       \item[iii.] the communication rate needed to send the index picked by the encoder $\mathcal{E}(S,W)$,
    \item[iv.] and the effect of the further precision $W_{\epsilon}$ on the communication cost, generalization error, and empirical risk.
\end{itemize}
Note that by studying the generalization error and the empirical risk, we indirectly provide an analysis of the population risk of the model $\mathcal{D}(K,W_{\epsilon})$.

\section{In-expectation performance}
In this section, we propose a source encoder based on the Ordered Random Coding (ORC) method and provide an in-expectation guarantee for the performance of the joint training and source coding scheme. ORC is a variant of MRC, with ``near-optimal'' in-expectation communication cost, that is proposed in \cite{theis2022algorithms} to overcome the communication cost of MRC, which grows with $N$. However, applying the ORC scheme directly to our problem setup would require the knowledge of $P_{W|S}$. Moreover, for deterministic learning algorithms $\mathcal{A}(S)$, the induced distribution $P_{W|S}$ is degenerate. As a result, both ``vanilla'' MRC and ORC fail in our case. 

To overcome this shortcoming, here, we consider an ORC-based encoder that uses an \emph{arbitrary quantization rule} $P_{\hat{W}|W}$ that can be chosen by the client, instead of $P_{W|S}$ which is induced by $\mathcal{A}(S)$. The performance of this modified ORC-based source encoder jointly with the learning algorithm $\mathcal{A}(S)$ is investigated in the following result.

\begin{theorem} \label{th:expectation} Suppose that the learning algorithm $\mathcal{A}(S)$ induces $P_{W|S}$. Suppose the loss function is $\mathfrak{L}$-Lipschitz, \ie $|\ell(z,w)-\ell(z,w')|\leq \mathfrak{L} \|w-w'\|$ for all $w,w'\in \mathcal{W}$ and $z \in \mathcal{Z}$. Consider a quantization set $\mathcal{\hat{W}}\subseteq \mathcal{W}$. Then, for any prior $Q$ defined over $\mathcal{\hat{W}}$, there exists a proper source encoder $\mathcal{E}(S,W)=K$, agnostic to $P_{W|S}$, such that for every $t>0$ the following conditions hold simultaneously.

\textbf{i. [Empirical risk]}  For every $(S,W)$, with probability at least $1-2\sqrt{b_W}$, 
\begin{align}
     \mathbb{E}_{K}&\Big[\hat{\mathcal{L}}(S,\mathcal{D}(K,W_{\epsilon}))\Big] \leq \mathbb{E}_{K}\big[\hat{\mathcal{L}}(S,W)\big]+  \\& \frac{2\mathfrak{L}\sqrt{ \mathbb{E}_{\hat{W}\sim P_{\hat{W}|W}}\left[\|W-\hat{W}\|^2\right]  b_W}-\mathfrak{L}\mathbb{E}_K[\Delta_{\vc{U}}(W,K)]}{1-\sqrt{b_W}},\nonumber
\end{align}    
where $\mathbb{E}_K[\cdot]$ denotes the expectation with respect to the stochasticity of the encoder and
\begin{align*}
    \hspace{-0.2 cm} b_W{\coloneqq} e^{-t/4}{+}2\sqrt{\mathbb{P}_{\hat{W}}\left({\log} \frac{\der P_{\hat{W}|W}}{\der Q}{>}D_{KL}(P_{\hat{W}|W}\|Q)+t/2\right)},
\end{align*}
where $\der P_{\hat{W}|W}/\der Q_{\hat{W}}$ is the Radon–Nikodym derivative of $P_{\hat{W}|W}$ with respect to $Q_{\hat{W}}$ and depends on $W$.

 \textbf{ii. [Generalization error]} With probability at least $1-\delta$,\footnote{Note that $D_{KL}\big(P_{\hat{W}|W}\|Q\big) = \mathbb{E}_{\hat{W}\sim P_{\hat{W}|W}} \big[{\log} (\der P_{\hat{W}|W}/\der Q_{\hat{W})}\big]$ is a random variable that depends on $W$.}
			\begin{align}
	\mathbb{E}_{K}\big[\gen&\left(S,\mathcal{D}(K,W_{\epsilon})\right)\big] \leq\nonumber \\
   &\sqrt{\frac{D_{KL}(P_{\hat{W}|W}\|Q)+t+\log(\sqrt{2n}/\delta)}{2n-1}+\varepsilon}, \label{eq:genDecoded}
			\end{align}
where
\begin{align*}
\varepsilon \leq & \inf_{\alpha_W} \sup_{\nu_W \in \mathcal{G}_{\delta}(W)} \big\{ \mathbb{E}_{W\sim \nu_W}\left[\mathbb{P}\left(\mathbb{E}_K[\|W_{\epsilon}\|] > \alpha_W\right)+4\mathfrak{L}\alpha_W \right]\big\} \\
    \leq & \varepsilon_{\delta}.
\end{align*}
Here, $\mathcal{G}_{\delta}(W)$ is the set containing all distributions $\nu_W$ over $\mathcal{W}$ such that $D_{KL}\left(\nu_W\|P_W\right) \leq \log(1/\delta)$ and
\begin{align*}
& \varepsilon_{\delta} {\coloneqq} \sup_{\nu_W \in \mathcal{G}_{\delta}(W)}
  \Big\{ 2\mathbb{E}_{W\sim \nu_W}\left[b_W\right] {+}8\sqrt{\mathfrak{L}\varepsilon_{\nu_W}\mathbb{E}_{W\sim \nu_W}\left[b_W\right]}\Big\}, \\
&   \varepsilon_{\nu_W} \coloneqq  \sqrt{\mathbb{E}_{W,\hat{W}\sim \nu_W P_{\hat{W}|W}}\left[\|W-\hat{W}\|^2\right]}.
\end{align*}

\textbf{iii. [Communication rate]} The expected communication cost (over $S$ and $W$)  of sending $\tilde{W}_{\vc{U}}[K]$ is no larger than
\begin{align*}
    C+\log(C+1)+4,
\end{align*}     
where $C\coloneqq \mathbb{E}_{W\sim P_W}\big[D_{KL}(P_{\hat{W}|W}\|Q)\big]$. 

Moreover, such source encoder can be constructed by ORC using the distributions $P_{\hat{W}|W}$ and $Q$.   
	\end{theorem}
 
 The theorem is proved in Section~\ref{pr:expectation}. Here, we make a few remarks about this result.  First, the bound on empirical risk in part i is composed of two terms. The first term is $\mathbb{E}_{K}[\hat{\mathcal{L}}(S,W)]$ and it can be made small at the transmitter side. The second term can be made small (note that $D_{KL}(P_{\hat{W}|W}\|Q)=\mathbb{E}[\log(\der P_{\hat{W}|W}/\der Q)]$) either by sending ``more precision'' (as discussed below) or by increasing $t$ which results in more computational complexity due to the increase of $N_W$ in \eqref{eq:orc} (see \cite{theis2022algorithms}).  However, increasing $t$ results in increasing the generalization bound and, as can be verified, in slightly increasing the \emph{optimal} communication rate.

Second, the bounds on the generalization error and the communication cost can be made small by minimizing the quantity $D_{KL}(P_{\hat{W}|W}\|Q)$. Hence, by minimizing this term one can guarantee both good generalization performance and a low communication rate. This shows that there exists an alignment between these two criteria. This finding suggests that using $D_{KL}(P_{\hat{W}|W}\|Q)$ as a regularizer jointly improves the generalizability and the needed communication rate. This is why in Fig~\ref{fig:fig1} we assumed that the learning algorithm $\mathcal{A}$ may have access to $Q$ which is used primarily for the source codebook generation. We note that a similar finding could be achieved for the \emph{recovered model} using the ORC encoder with respect to $P_{W|S}$ and some $Q_W$.

Third, in part i, it can be shown that as $\Delta_{\vc{U}}$ increases (equivalently as $\|W_{\epsilon}\|$ increases), the bound on the empirical risk decreases. Hence, sending further precision will improve (reduce) the empirical risk bound. In contrast, from the bound on $\varepsilon$ in part ii., it can be verified that $\varepsilon=0$ for no-precision case, and in general sending more ``precision'' affects negatively the generalization bound. Moreover, increasing $\|W_{\epsilon}\|$ naturally results in more communication overhead. Hence, overall, more precision will benefit the empirical risk guarantee, while having a negative effect on the generalization bound and the communication cost.

Fourth, it is easy to show that similar to the proof of part ii, the following bound on $\mathbb{E}_{W\sim P_{W|S}}\mathbb{E}_{K}\left[\gen(S,W)\right]$ holds with probability at least $1-\delta$,

\begin{align}
				\sqrt{\frac{C_S+t_S+\log(\sqrt{2n}/\delta)}{2n-1}+\varepsilon_{S}}, \label{eq:genOriginalExpectation}
			\end{align}
where 
\begin{align*}
 C_S\coloneqq& \mathbb{E}_{W\sim P_{W|S}}\big[D_{KL}(P_{\hat{W}|W}\|Q)\big], \\  
 t_S\coloneqq & \min(t,\log(C_S+1)+4),\\
 \varepsilon_{S} \coloneqq &
   2\mathbb{E}_{W\sim P_{W|S}}\left[b_W\right] +8\sqrt{\mathfrak{L}\varepsilon_{P_{W|S}}\mathbb{E}_{W\sim P_{W|S}}\left[b_W\right]}.
\end{align*}
A similar bound can be established when the expectation with respect to $(S,W)\sim P_{S,W}$ is considered. Furthermore, it can be verified that similarly to part iii, for a given $S$, the expected communication rate of $C_S+\log(C_S+1)+4$ can be achieved.

Finally, in our scheme, ORC is performed using the densities $P_{\hat{W}|W}$ and $Q$ in a way that is agnostic to $P_{W|S}$. This is a more practical result than the result of, \eg \cite{isik2023communication}, in which probability distribution $P_{W|S}$ induced by the learning algorithm should be known. Furthermore, using ORC allowed us to establish part iii. on the communication rate. A similar result cannot be established using the scheme of \cite{isik2023communication}, as they use MRC \cite{havasi2018minimal} in which the communication complexity grows linearly with $t$. The reader is referred to \cite{theis2022algorithms} for more details.

\section{One-shot performance}
The scheme proposed in the previous section provides an ``in average'' guarantee for the empirical risk and the generalization error of the recovered model at the receiving side. This is a relevant criterion mainly for multi-round schemes like federated learning. However, this may be insufficient for the one-shot scenario which is investigated in this section. More precisely, we study here the performance of a ``vector quantizer encoder'' defined as follows. Let $N$ be a fixed integer and
\begin{align*}
    \mathcal{E}_{\mathrm{VQ}}(S,W) = \argmin\nolimits_{n \in [N]} \big\|W-\tilde{W}_{\vc{U}}[n]\big\|.
\end{align*} 

For this encoder, the following result holds.

\begin{theorem} \label{th:one-shot} Consider the setup of Theorem~1. Fix some $\epsilon>0$. Let $K=\mathcal{E}_{\mathrm{VQ}}(S,W)$.

\textbf{i. [Empirical risk]} With probability at least $1-\tau_{\epsilon}$,
\begin{align}
     \hat{\mathcal{L}}(S,\mathcal{D}(&K,W_{\epsilon})) \leq \hat{\mathcal{L}}(S,W)+ 2 \mathfrak{L}\,\left(\epsilon-\Delta_{\vc{U}}(W,K)\right),\label{eq:distOneShot}
\end{align} 
where $\tau_{\epsilon}$ is defined as 
\begin{align*}
		\inf \Bigg\{\mathbb{E}_W\Big[\mathbb{P}\Big(\big(W,\hat{W}\big)&\notin \mathcal{F}_{\epsilon} \big| W\Big)^{N_2}\Big]+N_2\,e^{-\exp(\gamma)}\\&+ N_2\mathbb{P}_{W,\hat{W}}\left((W,\hat{W})\notin \mathcal{I}_{N_1,\gamma}\right)\Bigg\}.
	\end{align*}
In this definition, 
\begin{align*}
    \mathcal{I}_{N _1,\gamma} \coloneqq & \left\{(w,\hat{w})\colon \frac{\der P_{\hat{W}|W=w}}{\der Q_{\hat{W}}} (\hat{w}) \leq \log(N_1)-\gamma \right\},\\
   \mathcal{F}_{\epsilon} \coloneqq &\left\{(w,\hat{w})\colon \|w-\hat{w}\|\leq \epsilon\right\},
\end{align*}
and the infimum is over all $\gamma >0$, all Markov Kernels $P_{\hat{W}|W}$, and all $N_1,N_2$ that satisfy the following conditions:
\begin{itemize}[leftmargin=0.6 cm,topsep=1pt]
    \item[\textbf{a)}] $N_1 \times N_2 \leq N$, 
\item[\textbf{b)}] For all $W$, $1< \lambda^{N_2}+N_2 (1-\lambda)$ holds for $\lambda =\mathbb{P}\big(\big(W,\hat{W}\big)\notin \mathcal{F}_{\epsilon}\big|W\big)$, 
\item[\textbf{c)}]  $\mathbb{P}\big(\big(W,\hat{W}\big)\notin \mathcal{F}_{\epsilon} \medcap \mathcal{I}_{N_1,\gamma} \big|W\big) +e^{-\exp(-\gamma)}\leq 1$.
\end{itemize}

\textbf{ii. [Generalization error]}  With probability at least $1-\delta-\tau_{\epsilon}$,
    \begin{align}
\gen\left(S,\mathcal{D}(K,W_{\epsilon})\right) \leq \sqrt{\frac{\log(N)+\log(1/\delta)}{2n}} +2\mathfrak{L}\,\epsilon. \label{eq:genOriginalGreedy}
			\end{align}
 \end{theorem}
The theorem is proved in Section~\ref{pr:one-shot}.  
The significance of this theorem is that it does not consider the expectation with respect to the encoder, in contrast to Theorem~\ref{th:expectation}. It should be further noted that the scheme is agnostic to the distribution $P_{\hat{W}|W}$ which appears only in the ``failure probability'' analysis. 

\vspace{-0.1 cm}
\section{Proof of Theorem~\ref{th:expectation}} \label{pr:expectation}
\begin{IEEEproof} 
 Denote \vspace{-0.2 cm}
		\begin{align}
			\vspace{-0.4 cm}\rho_W(\hat{W}) \coloneqq \frac{\der P_{\hat{W}|W}}{\der Q_{\hat{W}}},
		\end{align}
where for a better clarity, $Q$ is denoted by $Q_{\hat{W}}$. Furthermore fix some $t>0$ and for any $w\in \mathcal{W}$, define
\begin{align*}
N_w\coloneqq  e^{L_w+t},\quad \quad L_w \coloneqq D_{KL}\big(P_{\hat{W}|W=w}\| Q_{\hat{W}}\big).
\end{align*}
We start by defining the stochastic source encoder $\mathcal{E}(S,W)$. Fix some $(W,\vc{U})$. Let $G_n$, $n\in[N_W]$ be i.i.d. instances from the Gumbel distribution \cite{gumbel1954statistical} with scale 1. Denote their ordered sequence by $\tilde{G}_1,\ldots,\tilde{G}_{N_W}$, \ie $\tilde{G}_1 \geq \cdots \geq \tilde{G}_{N_W}$. We define the encoder using the ORC method introduced in \cite{theis2022algorithms}. That is, $\mathcal{E}(S,W)=K$ is chosen according to the following rule:
\begin{align}
    K = \argmax_{n\leq N_W} \left\{\log\rho_W\left(\tilde{W}_{\vc{U}}[n]\right)+\tilde{G}_n\right\}. \label{eq:orc}
\end{align}

Now, we analyze the performance of this encoder.

\textbf{Part i.} Similar to the related proofs in \cite{havasi2018minimal,theis2022algorithms,isik2023communication}, this proof is also inspired by the ideas introduced in \cite{chatterjee2018sample}. 

Using the definition of $\Delta_{\vc{U}}(W,K)$ and the Lipschitz continuity assumption, to prove this part, it is sufficient to show that for every $(S,W)$, with probability at least $1-2\sqrt{b_W}$,
\begin{align*}
    \left\|W-\tilde{W}_{\vc{U}}[K]\right\| {\leq} \frac{2\sqrt{ \mathbb{E}_{\hat{W}\sim P_{\hat{W}|W}}\left[\|W-\hat{W}\|^2\right]  b_W}}{1{-}\sqrt{b_W}}.
\end{align*}
Due to \cite[Theorem~3.1.]{theis2022algorithms}, the distribution of $\tilde{W}_{\vc{U}}[K]$ is the same as the one picked using MRC introduced in \cite{havasi2018minimal,Cuff09}. Hence, for this section, we consider the following MRC encoder: Given $(W,\vc{U})$, let $\mathcal{E}(S,W)$ pick the index $i\in[N_W]$, with probability $\frac{\rho_W\left(\tilde{W}_{\vc{U}}[i]\right)}{\sum_{j\in[N_W]}\rho_W\left(\tilde{W}_{\vc{U}}[j]\right)}$.  Define
\begin{align}
\mathfrak{I}(W) \coloneqq & \mathbb{E}_{\hat{W} \sim P_{\hat{W}|W}}\left[\|W-\hat{W}\|\right]
\nonumber \\
= & \mathbb{E}_{\vc{U}}\left[\frac{1}{N}\sum_{i\in[N]} \|W-\tilde{W}_{\vc{U}}[i]\| \rho_W(\tilde{W}_{\vc{U}}[i])\right], \nonumber
\end{align}
where the equality comes from the fact that for every $i\in[N]$, $\tilde{W}_{\vc{U}}[i] \sim Q_{\hat{W}}$. Furthermore, let
\begin{align}
\mathfrak{I}_N(\vc{U},W) \coloneqq &\frac{1}{N}\sum_{i\in[N]} \|W-\tilde{W}_{\vc{U}}[i]\| \rho_W(\tilde{W}_{\vc{U}}[i]), \nonumber 
\end{align}
The dependence of $\mathfrak{I}_N(\vc{U},W)$ on $\vc{U}$ shows the dependence on a given draw of the codebook, using the common randomness $\vc{U}$. We first claim that for each $w$,
\begin{align}
			\mathbb{E}_{\vc{U}}&\big[\big| \mathfrak{I}_N(\vc{U},w)-\mathfrak{I}(w) \big|\big] \leq \sigma_{0,w} b_w \label{eq:claim1}, 
		\end{align}
		where $\sigma_{0,w}^2 \coloneqq \mathbb{E}_{\hat{W}\sim P_{\hat{W}|W=w}}\big[\|w-\hat{W}\|^2\big]$.
 This claim can be proved using \cite[Theorem~1.1]{chatterjee2018sample}, as shown in Appendix~\ref{app:claim}. 

Now, due to the choice of the encoder, we have 
\begin{align}
   \mathbb{E}_{K}\left[\|W-\tilde{W}_{\vc{U}}[K]\|\right]= &\sum_{i\in[N]}  \frac{\|W-\tilde{W}_{\vc{U}}[i]\|\rho_W(\tilde{W}_{\vc{U}}[i]) }{\sum_{i\in[N]} \rho_W(\tilde{W}_{\vc{U}}[i])}\nonumber \\
    =& \frac{\mathfrak{I}_N(\vc{U},W)}{\frac{1}{N}\sum_{i\in[N]} \rho_W(\tilde{W}_{\vc{U}}[i])}.
\end{align}
Then, from one hand, for any $\epsilon_w>0$, \eqref{eq:claim1} concludes that
\begin{align}
    \mathbb{P}\left(\left|\mathfrak{I}_N(\vc{U},w)-\mathfrak{I}(w)\right|> \tau_w\right) \leq \sigma_{0,w} b_w/\tau_w,
\end{align}
and from the other hand, due to \cite{chatterjee2018sample}, 
\begin{align}
    \mathbb{P}\Big(\Big|\frac{1}{N}\sum\nolimits_{i\in[N]} \rho_w(\tilde{W}_{\vc{U}}[i]) -1 \Big|> \epsilon_w\Big)\leq b_w/\epsilon_w.
\end{align}
Moreover, whenever $\left|\mathfrak{I}_N(\vc{U},w)-\mathfrak{I}(w)\right|\leq \tau_w$ and $\|\frac{1}{N}\sum_{i\in[N]} \rho_w(\tilde{W}_{\vc{U}}[i]) -1 \|\leq \epsilon_w$,  we have
\begin{align}
     \mathbb{E}_{K}\left[\|w-\tilde{W}_{\vc{U}}[K]\|\right] \leq \frac{\tau_w+\mathfrak{I}(w)\epsilon_w}{1-\epsilon_w}.
\end{align}
Let $\tau_w=\sigma_{0,w}\epsilon_w$. Then, for every $w$, with probability at least $1-2b_w/\epsilon_w$, we have
\begin{align}
     \mathbb{E}_{K}\left[\|w-\tilde{W}_{\vc{U}}[K]\|\right] \leq \frac{2\sigma_{0,w}\epsilon_w}{1-\epsilon_w}. \label{eq:distORC}
\end{align}
Letting $\epsilon_w=\sqrt{b_w}$ completes the proof of this part.

\textbf{Part ii.} The generalization bound can be established using \cite[Theorem~1]{sefidgaran2023datadependent}. To show this, it is sufficient to show that the learning algorithm $\tilde{A}\colon \mathcal{Z}^n \to \hat{\mathcal{W}}$, defined as 
\begin{align*}
    \tilde{A}(S) = \mathcal{D}\left(K,W_{\epsilon}\right)=\tilde{W}_{\vc{U}}\left[\mathcal{E}\left(S,\mathcal{A}(S)\right)\right]+W_{\epsilon},
\end{align*}
is $(\log(N_{W}),\varepsilon_,\delta;d_m)$-compressible in the sense defined in \cite[Definition~2]{sefidgaran2023datadependent}. The condition on the rate is trivial due to the way the codebook $\mathcal{H}_{\vc{U},N}$ is constructed and the encoder is defined. It remains to bound the distortion constraint
\begin{align*}
     \varepsilon {=}\sup_{\nu_W\in\mathcal{G}_{\delta}(W)} \mathbb{E}_{W,K}\left[\gen(S,\mathcal{D}\left(K,W_{\epsilon}\right))^2{-}\gen(S,\tilde{W}_{\vc{U}}[K])^2\right].
\end{align*}
Now, for any $\alpha_W$ and any distribution $\nu_W \in \mathcal{G}_{\delta}(W)$, we have
\begin{align*}
&\hspace{-0.4 cm}\varepsilon_{\alpha_W,\nu_W} \coloneqq\mathbb{E}_{W,K}\left[\gen(S,\mathcal{D}\left(K,W_{\epsilon}\right))^2-\gen(S,\tilde{W}_{\vc{U}}[K])^2\right]\\
\stackrel{(a)}{\leq} & \mathbb{E}_{W}\left[\mathbb{P}\left(\mathbb{E}_K\|W_{\epsilon}\| > \alpha_W \right)\right]+2\times \\
&\hspace{-0.2 cm} \mathbb{E}_{W,K}{\Big[}\big|\gen(S,\mathcal{D}\left(K,W_{\epsilon}\right)){-}\gen(S,\tilde{W}_{\vc{U}}[K])\big|\Big| \mathbb{E}\|W_{\epsilon}\| {\leq} \alpha_W {\Big]}\\\stackrel{(b)}{\leq}  & \mathbb{E}_{W}\left[\mathbb{P}\left(\mathbb{E}_K\|W_{\epsilon}\| > \alpha_W \right)\right]+4\mathfrak{L}\mathbb{E}_{W}[\alpha_W],
\end{align*}
where $W\sim \nu_W$, $(a)$ is concluded since $\ell(z,w) \in [0,1]$, and $(b)$ is derived by Lipschitzness and since $W_{\epsilon}=\mathcal{D}\left(K,W_{\epsilon}\right)-\tilde{W}_{\vc{U}}[K]$. Now, combining the above inequality with
\begin{align*}
    \varepsilon \leq \inf\nolimits_{\alpha_W} \sup\nolimits_{\nu_W \in \mathcal{G}_{\delta}(W)} \varepsilon_{\alpha_W,\nu_W},
\end{align*}
proves the first bound on $\varepsilon$.

Next, note that  by assumption $\|W_{\epsilon}\| \leq \|W-\tilde{W}_{\vc{U}}[K]\|$. Hence, $\varepsilon_{\alpha_W,\nu_W}$ is upper bounded further by 
\begin{align*}
\mathbb{E}_{W}\left[\mathbb{P}\left(\mathbb{E}_K\big[\|W-\tilde{W}_{\vc{U}}[K](U)\|\big] {>} \alpha_W \right)\right]{+}4\mathfrak{L}\mathbb{E}_{W}[\alpha_W].
\end{align*}
Now, using \eqref{eq:distORC} by letting $\alpha_W \mapsto \frac{2\sigma_{0,W}\epsilon_W}{1-\epsilon_W}$ and $\epsilon_W = \frac{\sqrt{b_W}}{\sqrt{b_W}+\sqrt{4\mathfrak{L}\sigma_{0,W}}}$, we have 
\begin{align*}
\varepsilon_{\alpha_W,\nu_W} \leq& 2\mathbb{E}_{W}\left[b_W/\epsilon_W\right] +8\mathfrak{L}\,\mathbb{E}_{W}\left[\frac{\sigma_{0,W}\epsilon_W}{1-\epsilon_W}\right] \\
=&2\mathbb{E}_{W}\left[b_W\right] +8\sqrt{\mathfrak{L}}\mathbb{E}_{W}\left[\sqrt{b_W \sigma_{0,W}}\right]\\
    \leq &2\mathbb{E}_{W}\left[b_W\right] +8\sqrt{\mathfrak{L}} \sqrt{\mathbb{E}_{W}\left[b_W\right] \mathbb{E}_W\left[\sigma_{0,W}\right]}\\
    \leq & 2\mathbb{E}_{W}\left[b_W\right] +8\sqrt{\mathfrak{L}\varepsilon_{\nu_W} \mathbb{E}_{W}\left[b_W\right]} .
\end{align*}
This concludes that $ \varepsilon \leq \varepsilon_{\delta}$, which completes the proof. 

\textbf{Part iii.} This part can be concluded similar to \cite[Corollary~3.4.]{theis2022algorithms}. The only difference with \cite[Corollary~3.4.]{theis2022algorithms} is that here, we allow different sizes of the codebook $N_W$. The proof however remains the same using \cite[Theorem~3.3.]{theis2022algorithms}.
\end{IEEEproof}

\section{Proof of Theorem~\ref{th:one-shot}} \label{pr:one-shot}
 \begin{IEEEproof}
\textbf{Part i.} Using the Lipschitz continuity assumption and the definition of $\Delta_{\vc{U}}[K,W]$, it suffices to show that $ \mathbb{P}\big(\big\| W-\tilde{W}_{\vc{U}}[K] \big\| > \epsilon \big) \leq \tau_{\epsilon}$. Note that
\begin{align}
    \mathbb{P}&\big(\big\| W-\tilde{W}_{\vc{U}}[K] \big\| > \epsilon \big) =\mathbb{P}\big(\forall i\in[N] \colon \big(W,\tilde{W}_{\vc{U}}[i]\big)\notin \mathcal{F}_{\epsilon} \big). \nonumber 
\end{align}
For simplicity denote $\mathcal{I}\coloneqq \mathcal{I}_{N_1,\gamma}$. Now, first using the proof of \cite[Lemma~4]{verdu2015non}, for any $j\in[N_2]$ and a given $W$, we have
 \begin{align}
   \mathbb{P}\Big(\forall i\in&\mathcal{N}_j \colon \big(W,\tilde{W}_{\vc{U}}[i]\big)\notin \mathcal{F}_{\epsilon} \big|W \Big)\nonumber \\
   \leq & \mathbb{P}\left(\big(W,\hat{W}\big)\notin \mathcal{F}_{\epsilon}\medcap \mathcal{I} \big|W\right) +e^{-\exp(-\gamma)},
\end{align}
where $\mathcal{N}_j=\{(j-1)N_2+1,\ldots,(j-1)N_2+N_1\}$. Hence,
\begin{align}
   \mathbb{P}\Big(\forall i\in&[N] \colon \big(W,\tilde{W}_{\vc{U}}[i]\big)\notin \mathcal{F}_{\epsilon} \big|W \Big)\nonumber \\
  \leq & \left( \mathbb{P}\left(\big(W,\hat{W}\big)\notin \mathcal{F}_{\epsilon}\medcap \mathcal{I} \big|W\right) +e^{-\exp(-\gamma)}\right)^{N_2} \nonumber \\
 = & \Big( \mathbb{P}\left(\big(W,\hat{W}\big)\notin \mathcal{F}_{\epsilon} \big|W\right)+\mathbb{P}\left(\big(W,\hat{W}\big)\in \mathcal{F}_{\epsilon} \setminus \mathcal{I} \big|W\right) \nonumber \\
  &\hspace{0.3 cm}+e^{-\exp(-\gamma)}\Big)^{N_2} \nonumber \\
    \stackrel{(a)}{\leq} &  \mathbb{P}\left(\big(W,\hat{W}\big)\notin \mathcal{F}_{\epsilon}\big|W\right)^{N_2}\nonumber \\
   &+N_2 \left( \mathbb{P}\left(\big(W,\hat{W}\big)\notin \mathcal{I} \big|W\right) +e^{-\exp(-\gamma)}\right),\label{eq:proofOneshot1}
\end{align}
where the last step holds using the union bound, the conditions of the theorem, and due to the inequality $(x+y)^n \leq x^n+ny$, that holds whenever $x,y \geq 0$, $x+y \leq 1$, and $1 \leq x^n+n(1-x)$.

Finally averaging with respect to $W$ yields the result.

\textbf{Part ii.} The proof of this part follows from the Lipschitz assumption, the fact that $\ell(z,w)\in[0,1]$, and part i, similar to part ii of Theorem~\ref{th:expectation}.
\end{IEEEproof}

\newpage
\bibliographystyle{IEEEtran}
% \enlargethispage{-1.2cm} 
% \IEEEtriggeratref{17}
\bibliography{biblio}

\appendices

\section{Proof of the claim \texorpdfstring{\eqref{eq:claim1}}{(12)}} \label{app:claim}
In this appendix, we prove the claim \eqref{eq:claim1}. Fix some $w\in \mathcal{W}$ and for ease of notations, let 
\begin{align*}
    a \coloneqq e^{L_w+t/2}.
\end{align*}
Define $\|w-x\|_{-a}$ as follows: $\|w-x\|_{-a} =0$, whenever $\rho_w(x)> a$, and equals $\|w-x\|$ otherwise.  Now, we can bound $\mathbb{E}_{\vc{U}}\left[\left| \mathfrak{I}_N(\vc{U},w)-\mathfrak{I}(w) \right|\right]$ as following:
\begin{align}
&\mathbb{E}_{\vc{U}}\left[\left| \mathfrak{I}_N(\vc{U},w)-\mathfrak{I}(w) \right|\right] \leq \nonumber \\  &\mathbb{E}_{\vc{U}}\left[\left| \mathfrak{I}_N(\vc{U},w)-\frac{1}{N}\sum_{i\in[N]} \|w-\tilde{W}_{\vc{U}}[i]\|_{-a} \rho_w(\tilde{W}_{\vc{U}}[i])\right|\right] \label{eq:pr1} \\
&+ \left| \mathbb{E}_{\hat{W}\sim P_{\hat{W}|W=w}}\left[\|w-\hat{W}\|_{-a}\right]-\mathfrak{I}(w) \right| \label{eq:pr2}  \\
&+\mathbb{E}_{\vc{U}}\left[\left| B \right|\right], \label{eq:pr3} 
\end{align} 
where  in \eqref{eq:pr3},
\begin{align}
    B \coloneqq & \frac{1}{N}\sum_{i\in[N]} \|w-\tilde{W}_{\vc{U}}[i]\|_{-a} \rho_w(\tilde{W}_{\vc{U}}[i]) \nonumber\\
    &-\mathbb{E}_{\hat{W}\sim P_{\hat{W}|W=w}}\left[\|w-\hat{W}\|_{-a}\right].
\end{align}
  
The term \eqref{eq:pr1} is bounded by
		\begin{align}
			\mathbb{E}_{\vc{U}}&\left[\left| \mathfrak{I}_N(\vc{U},w)-\frac{1}{N}\sum_{i\in[N]} \|w-\tilde{W}_{\vc{U}}[i]\|_{-a} \rho_w(\tilde{W}_{\vc{U}}[i])  \right|\right] \nonumber\\
			\stackrel{(a)}{\leq} & \mathbb{E}_{\vc{U}}\left[\mathbbm{1}_{\{\rho_w(\tilde{W}_{\vc{U}}[1])>a\}} \|w-\tilde{W}_{\vc{U}}[1]\| \rho_w(\tilde{W}_{\vc{U}}[1]) \right]  \nonumber \\
			= & \mathbb{E}_{\hat{W}\sim  P_{\hat{W}|W=w}}\left[\mathbbm{1}_{\{\rho_w(\hat{W})>a\}} \|w-\hat{W}\|  \right]  \nonumber\\
			\stackrel{(b)}{\leq} & \sqrt{\mathbb{E}_{\hat{W}\sim  P_{\hat{W}|W=w}}\left[\|w-\hat{W}\|^2\right] \mathbb{P}_{\hat{W}}\left(\rho_w(\hat{W})>a\right)}\nonumber \\
			=& \sigma_{0,w} \sqrt{\mathbb{P}_{\hat{W}}\left(\rho_w(\hat{W})>a\right)},
		\end{align}
  where $(a)$ is derived using the definition of $I_{n}(\vc{U},w)$ and since $\tilde{W}_{\vc{U}}[i]$ are generated in an i.i.d. manner and $(b)$ is derived using the Cauchy–Schwarz inequality.

Similarly \eqref{eq:pr2} can be upper bounded as
		\begin{align}
			\mathbb{E}_{\vc{U}}\Big[\big| \mathbb{E}_{\hat{W}\sim P_{\hat{W}|W=w}}&\left[\|w-\hat{W}\|_{-a}\right]-I \big|\Big] \nonumber \\
    &\leq \sigma_{0,w} \sqrt{\mathbb{P}_{\hat{W}}\left(\rho_w(\hat{W})>a\right)}.
		\end{align}

  Finally the last term \eqref{eq:pr3} squared can be upper bounded as
		\begin{align}
			\mathbb{E}_{\vc{U}}\left[\left| B \right|\right]^2 \leq & \text{Var}\big(\frac{1}{N}\sum_{i\in[N]} \|w-\tilde{W}_{\vc{U}}[i]\|_{-a} \rho_w(\tilde{W}_{\vc{U}}[i]) \big) \nonumber \\
   \leq & \text{Var}\left(\|w-\tilde{W}_{\vc{U}}[1]\|_{-a} \rho_w(\tilde{W}_{\vc{U}}[1])\big]\right)/N\nonumber \\ 
    \leq & \mathbb{E}_{\vc{U}}\left[ \|w-\tilde{W}_{\vc{U}}[1]\|_{-a}^2 \rho_w(\tilde{W}_{\vc{U}}[1])^2\big]\right] /N \nonumber \\
    \leq & a \mathbb{E}_{\vc{U}}\left[ \|w-\tilde{W}_{\vc{U}}[1]\|^2 \rho_w(\tilde{W}_{\vc{U}}[1])\right]/N \nonumber \\
    = & a \mathbb{E}_{\hat{W}}\left[ \|w-\hat{W}\|^2 \right]/N \nonumber \\
=&   a \sigma_0^2 /N.
		\end{align}
This completes the proof of \eqref{eq:claim1}.

\end{document}